\documentclass[10pt,conference]{IEEEtran}

\usepackage{amsfonts}
\usepackage{setspace}
\usepackage{afterpage}

\usepackage{setspace}
\usepackage{amssymb}
\usepackage{epsfig}
\usepackage{graphicx}
\usepackage{xcolor}
\usepackage{algorithm}
\usepackage{algorithmicx}
\usepackage{algpseudocode}
\usepackage{amsmath}

\floatname{algorithm}{Algorithm}

\usepackage[ps2pdf,
bookmarks=false,
bookmarksnumbered=false, 
bookmarksopen=false, 
colorlinks=false]{}

\begin{document}

\title{Deep Actor-Critic Reinforcement Learning for Anomaly Detection}

\author{Chen Zhong, M. Cenk Gursoy, and Senem Velipasalar
\\Department of Electrical Engineering and Computer Science,
Syracuse University, Syracuse, NY 13244
\\Email: czhong03@syr.edu, mcgursoy@syr.edu, svelipas@syr.edu}

\maketitle

\begin{abstract}
Anomaly detection is widely applied in a variety of domains, involving for instance, smart home systems, network traffic monitoring, IoT applications and sensor networks. In this paper, we study deep reinforcement learning based active sequential testing for anomaly detection. We assume that there is an unknown number of abnormal processes at a time and the agent can only check with one sensor in each sampling step. To maximize the confidence level of the decision and minimize the stopping time concurrently, we propose a deep actor-critic reinforcement learning framework that can dynamically select the sensor based on the posterior probabilities. We provide simulation results for both the training phase and testing phase, and compare the proposed framework with the Chernoff test in terms of claim delay and loss.
\end{abstract}


\section{Introduction}

Anomaly detection has been extensively studied in various fields, with applications in different domains. For instance, the authors in \cite{rajasegarar2008anomaly} provided a survey of anomaly detection techniques for wireless sensor networks. In \cite{kanev2017anomaly}, authors reviewed the problem of anomaly detection in home automation systems. In this paper, we specifically consider active hypothesis testing for the anomaly detection problem in which there are $k$ abnormal processes out of $N$ processes, where $0 \leq k \leq N$. During the detection process, the decision maker is allowed to observe only one of the $N$ processes at a time. The distribution of the observations depends on whether the target is normal or not. In this setting, the objective of the decision maker is to minimize the observation delay and dynamically determine all abnormal processes.

The original active hypothesis testing problem was investigated in \cite{chernoff1959sequential}. Based on this work, several recent studies proposed more advanced anomaly detection techniques in more complicated and realistic situations. For example, the authors in \cite{cecchi2017adaptive} considered the case where the decision maker has only limited information on the distribution of the observation under each hypothesis. In \cite{chen2019active}, the performance measure is the Bayes risk that takes into account not only the sample complexity and detection errors, but also the costs associated with switching across processes. Moreover, authors in \cite{cohen2015active} considered the scenario that in some of the experiments, the distributions of the observations under different hypotheses are not distinguishable, and extended this work to a case with heterogenous processes \cite{huang2019active}, where the observation in each cell is independent and identically distributed (i.i.d.). Also, the study of stopping rule has drawn much interest. For instance, in \cite{leonard2018robust}, improvements were achieved over prior studies since the proposed decision threshold can be applied in more general cases. The authors in \cite{zhang2018statistical} leveraged the central limit theorem for the empirical measure in the test statistic of the composite hypothesis Hoeffding test,  so as to establish weak convergence results for the test statistic, and, thereby, derive a new estimator for the threshold needed by the test.

Recently, machine learning-based methods have also been applied to such hypothesis testing problems. In \cite{kartik2018policy} and \cite{puzanov2018deep}, the deep Q-network has been employed for sequential hypothesis testing and change point detection, respectively, and in \cite{moustafa2019outlier} an adversarial statistical learning method has been proposed for anomaly detection. In this paper, we propose a deep actor-critic reinforcement learning framework to dynamically select the process to be observed and maximize the confidence level.


\section{System Model}

In this work, we consider $N$ independent processes, where each of the processes could be in either normal or abnormal state. We assume that at any time $t$, the probability of the process $i$, for $i = 1, 2, ..., N$, being abnormal is $P_i$. We denote the number of abnormal processes as $k$, and since all processes are assumed to be independent, the value of $k$ could be any integer in the range $[0, N]$ at any given time. It is also assumed that at any time instant, if anomaly occurs in any number of processes, the states of all processes will remain the same until all abnormal processes are detected and fixed.

We assume that there is a single observation target $Y_t$ for all processes, and the samples have different density distributions depending on the states of the processes (e.g., normal or abnormal).  For example, we can consider the scenario in which for each process, there is a sensor that can send a state signal to the observer in each time slot. When the process is normal, the sensor should send $Y = 0$, while if the process is abnormal, the sensor should send $Y = 1$.  We note that in practical settings the sensors are not always reliable, so in this work we assume that the sensor will erroneously send a flipped signal with probability $\rho$. Now, when the process is normal, the samples are distributed according to the Bernoulli distribution $Y \sim f(Y, \rho)$, and when the process is abnormal, the distribution of the samples follows the Bernoulli distribution $Y \sim g(Y, 1-\rho)$. Furthermore, we assume that the observer can only observe the sample from one of the $N$ sensors at any given time. Hence, to minimize the time slots needed for detecting the anomalies, it is important to find an effective policy for sensor selection.

Since there are $N$ processes, an unknown number of which can be in abnormal state, we have $M = 1 + \sum\limits_{k = 1}^{N} \binom{N}{k}$ hypotheses, where $k$ is the number of abnormal processes at a given time. We say $H_0 = \{\emptyset\}$ is true when none of the $N$ processes is abnormal. And for each of the $M-1$ possible combinations of unknown numbers of abnormal processes, we define a hypothesis $H_m$ for $m = 1, \dots, M-1$. Table \ref{table:model} shows the observation models along with the corresponding sample distribution at different sensors when the given hypothesis is true. In the table, we have three processes and we use $g$ and $f$ to denote the real sample density distributions in abnormal and normal states, respectively. For instance, hypothesis $H_4$ indicates that processes 1 and 2 are abnormal and hence the samples at sensors 1 and 2 follow the distribution $g$. On the other hand, samples in sensor 3 are distributed according to $f$ since process 3 is normal. It is important to note that we assume that the parameters of the sample density distributions are unknown to the observer. To obtain an approximation of the density distribution, we employ the maximum likelihood estimation. Here, we define $\Omega^{t}$ as the sample space at time $t$, which contains all samples $\{Y_1, Y_2, \dots, Y_t\}$. And $\mathcal{F}_{i,m}$ is a subset of $\Omega^{t}$, and it contains all samples collected from sensor $i$ when the hypothesis $H_m$ is true. And the estimated sample density distributions can be defined as $f(Y_t|\mathcal{F}_{i,m})$ and $g(Y_t|\mathcal{F}_{i,m})$.

\begin{table}[]
	\centering
	\caption{Observation Model}
	\label{table:model}
	\begin{tabular}{|c|c|c|c|}
		\hline
		& sensor 1 & sensor 2 & sensor 3 \\ \hline
		$H_0$ = \{$\emptyset$\} & f        & f        & f        \\ \hline
		$H_1$ = \{1\}                        & g        & f        & f        \\ \hline
		$H_2$ = \{2\}                        & f        & g        & f        \\ \hline
		$H_3$ = \{3\}                        & f        & f        & g        \\ \hline
		$H_4$ = \{1, 2\}                     & g        & g        & f        \\ \hline
		$H_5$ = \{1, 3\}                     & g        & f        & g        \\ \hline
		$H_6$ = \{2, 3\}                     & f        & g        & g        \\ \hline
		$H_7$ = \{1, 2, 3\}                  & g        & g        & g        \\ \hline
	\end{tabular}
\end{table}

We denote the prior probability of each hypothesis being true as $\pi = [\pi_0, \dots, \pi_{M-1}]$. Since the probability of the process $i$ being abnormal is assumed to be $P_i$, the prior probabilities are the joint probabilities of the $N$ processes being in the corresponding states. Then, we denote $\pi_m^t$ as the posterior belief of the hypothesis $H_m$ being true at time $t$, and the posterior belief is updated as

\begin{equation}\label{eq:updatepi}
\pi_{m}^{t} = \frac{\pi_m \prod\limits_{t = 1}^{T} p_{m}^{i_t}(Y_t|\mathcal{F}_{i_t, m})}{\sum\limits_{l = 0}^{M-1}\pi_l \prod\limits_{t = 1}^{T} p_{l}^{i_t}(Y_t|\mathcal{F}_{i_t,l})}
\end{equation}
where we denote the sensor selected at time $t$ by $i_t$, and
\begin{align} \label{eq:MU-SI}
 p_{m}^{i_t}(Y_t|\mathcal{F}_{i_t,m})=
\begin{cases}
g(Y_t|\mathcal{F}_{i,m}) \hspace{0.5cm} \text{if $i_t \in H_m$}\\
f(Y_t|\mathcal{F}_{i_t,m}) \hspace{0.5cm}  \text{if $i_t \notin H_m$}
\end{cases}.
\end{align}

\section{Problem Formulation}

Similar to \cite{kartik2018policy} and \cite{naghshvar2013active}, we consider the confidence level as the maximization objective. The confidence level on hypothesis $H_m$ being true is given by the Bayesian log-likelihood ratio $\mathcal{C}_{H_m}$:
\begin{equation}
\mathcal{C}_{H_m} := \log \frac{\pi_m}{1-\pi_m}.
\end{equation}
And the average Bayesian log-likelihood ratio is defined as
\begin{equation}
\mathcal{C} = \sum\limits_{m = 0}^{M-1} \pi_m \log \frac{\pi_m}{1 - \pi_m} = \sum\limits_{m = 0}^{M-1} \pi_m \mathcal{C}
_{H_m}.
\end{equation}

\begin{figure}
\centering
\includegraphics[width=0.95\linewidth]{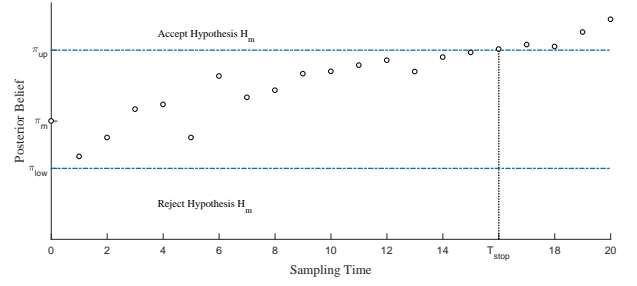}
\caption{An example of stopping time.}
\label{fig:PI}
\end{figure}

While maximizing the long term average confidence level, we also aim at minimizing the stopping time, $T_{\text{stop}}$. So we assume that there are upper bound and lower bound on the posterior belief. As shown in Fig.\ref{fig:PI}, the hypothesis $H_m$ is claimed to be accepted when the posterior belief $\pi_m$ is greater than the upper bound $\pi_{\text{up}}$, or to be rejected when the posterior belief is less than the lower bound $\pi_{\text{low}}$. And once any of the $M$ hypotheses is accepted, the observer stops receiving samples immediately.


\section{Deep Actor-Critic Framework}

In this section, we describe the proposed deep actor-critic learning framework for the anomaly detection problem.

\subsection{Preliminaries}
We first introduce the relevant definitions within the framework.

\emph{Agent's Observation and State:} Since the agent can only observe one sample $Y_t$ from the selected sensor $i_t$ at time $t$, the problem can be modeled as a partially observable Markov decision process (POMDP). With this sample, the agent can update the posterior belief $\pi^t$ according to (\ref{eq:updatepi}). And we take the posterior belief vector as the state (or input) of the agent, and we denote the state at time $t$ as $\mathcal{O}_t$, and define it as
\begin{equation}
 \mathcal{O}_t=
 \begin{cases}
 \pi \hspace{1.5cm} \text{t = 1}\\
 \pi^{t-1} \hspace{0.5cm}  \text{otherwise}
 \end{cases}.
\end{equation}

\emph{Action:} We denote the action space as $\mathcal{A}$, in which all valid actions are included. Here, the size of the action space is $N$, and a valid action stands for selecting the corresponding sensor and receiving the sample to update the posterior belief. In each iteration, the agent will score all valid actions, and choose the one with the highest score to execute.

\emph{Reward:} As we introduced in the previous sections, the proposed agent has two goals: 1) maximize the average confidence level and 2) minimize the stopping time. So we define the immediate reward $r_t$ as
\begin{equation}
r_t = \frac{\mathcal{C}^t - \mathcal{C} }{t},
\end{equation}
where $\mathcal{C}^t = \sum\limits_{m = 0}^{M-1} \pi_m^t \log \frac{\pi_m^t}{1 - \pi_m^t}$.

Here, we define the state $\mathcal{O}_T$ as the terminal state if any of the $M$ hypothesis is claimed to be accepted, i.e., $\max(\pi^{T-1}) \geq \pi_{\text{up}}$. And when we update the agent, we consider a weighted reward $R_t$ at time $t \leq T$, as a discounted sum of the rewards:
\begin{equation}
	R_t = \sum_{\tau = t}^{T} \lambda^{\tau - t} r_{\tau},
\end{equation}
so that each previous selection that can lead to better future steps will achieve a greater reward. And in the implementation, the agent will be updated $T$ times after the terminal state has been reached, using the weighted reward achieved at the terminal time $T$, and all the way back to the initial time $t = 0$.


\subsection{Algorithm Overview}\label{sub:algorithm}
In this subsection, we describe the architecture of the actor-critic algorithm. The actor-critic architecture consists of two neural networks: actor and critic. In our model, these two networks will not share any neurons but are parameterized by $\theta$.



\emph{Actor:} The actor is employed to explore a policy $\mu$ that maps the agent's observation $\mathcal{O}$ to the action space $\mathcal{A}$:
\begin{equation}
	\mu_{\theta}(\mathcal{O}) : \mathcal{O} \rightarrow \mathcal{A}.
\end{equation}
So the mapping policy $\mu_{\theta}(\mathcal{O})$ is a function of the observation $\mathcal{O}$ and is parameterized by $\theta$. And the chosen action can be denoted as
\begin{equation}
	a = \mu_{\theta}(\mathcal{O})
\end{equation}
where we have $a \in \mathcal{A}$. Since the action space is discrete, we use the softmax function at the output layer of the actor network so that we can obtain the scores of each actions. The scores sum up to $1$ and can be regarded as the probabilities of obtaining a good reward when the corresponding actions are chosen.

\emph{Critic:} The critic is employed to estimate the value function $V(\mathcal{O})$. At time instant $t$, when action $a_t$ is chosen by the actor network, the agent will execute it in the environment and send the current observation $\mathcal{O}_t$ along with the feedback from the environment to the critic. The feedback includes the reward $r_t$ and the next time instant observation $\mathcal{O}_{t+1}$. Then, the critic calculates the TD (Temporal Difference) error:
\begin{equation}
	\delta^{\mu_\theta} = r_t + \gamma V(\mathcal{O}_{t+1}) - V(\mathcal{O}_t)
\end{equation}
where $\gamma \in (0,1)$ is the discount factor.

\emph{Update:} The critic is updated by minimizing the least squares temporal difference (LSTD):
\begin{equation}
	V^* = \arg \min_{V} (\delta^{\mu_\theta} )^2
\end{equation}
where $V^*$ denotes the optimal value function.

The actor is updated by policy gradient. Here, we use the TD error to compute the policy gradient:
\begin{equation}
	\nabla_{\theta} J(\theta) = E_{\mu_{\theta} } [ \nabla_{\theta} \log \mu_\theta(\mathcal{O}, a)  \delta^{\mu_\theta} ]
\end{equation}
where $\mu_\theta(\mathcal{O}, a)$ denotes the score of action $a$ under the current policy. Then, the weighted difference of parameters in the actor at time $t$ can be denoted as $\Delta\theta_{t} = \alpha \nabla_{\theta_t} \log \mu_{\theta_t}(\mathcal{O}_t, a_t) \delta^{\mu_{\theta_t} }$, where $\alpha \in (0,1)$ is the learning rate. And the actor network $i$ can be updated using the gradient descent method:
\begin{equation}
	\theta_{t+1} = \theta_t + \alpha \nabla_{\theta_t} \log \mu_{\theta_t}(\mathcal{O}_t, a_t) \delta^{\mu_{\theta_t} }.
\end{equation}

\subsection{Training Phase}
In the training phase, the actor and critic neural networks are constructed and trained. For each episode, there will be a true hypothesis, generated according to the prior belief $\pi$. The agent will observe one sample at a time until it can accept a hypothesis. In the episode, at the beginning of each time slot $t$, the agent receives the current state $\mathcal{O}_t$, and chooses one out of the $N$ sensors to obtain a sample $Y_t$. Based on the sample, the agent can update the posterior belief $\pi^{t}$ and receive a reward. Then the critic network and actor network will be updated. Since the agent does not know which hypothesis is indeed true, the samples will be added to the corresponding subsets of overall sample space after the ground-truth is revealed, i.e., the posterior belief is always updated by the estimated density distribution based on the samples collected in the previous episodes.

The full framework is provided in Algorithm \ref{alg:AC-Train} below on the next page.

\begin{algorithm}
	\caption{Deep Actor-Critic Reinforcement Learning Algorithm for Anomaly Detection: Training Phase}
	\label{alg:AC-Train}
	\begin{algorithmic}
		\State $t = 0$
		\State Initialize the critic network $V_{\theta}(\mathcal{O} )$ and the actor $\mu_{\theta}(\mathcal{O})$, parameterized by $\theta$.
		
		\State The agent initializes the sample space $\Omega^0$, and the subsets $\mathcal{F}_{i,m}$, for $i = 1,\dots, N$ and $m = 0, \dots, M-1$.
		\For{$T = 1 : \text{Maximum episode}$}
		\State $t_{\text{start}} = t$
		\State Generate a new hypothesis $H_j$ to be true according to the prior belief $\pi$, and $j \in \{0,1, \dots, M-1\}$.
		\State The agent fetches the prior belief vector $\pi$ as the initial state.
		\While{$\mathcal{O}_T$ is not a terminal state}
		\State $t \leftarrow t+1$
		
		\State With the state $\mathcal{O}_t$, the agent selects one out of the $N$ sensors according to the decision policy $a_t = \mu(\mathcal{O}_t| \theta)$ w.r.t. the current policy.
		\State Agent receives the sample $Y_t$ from the chosen sensor and update the posterior belief vector $\pi^t$.
		\State Agent updates the sample $Y_t$ to the sample space $\Omega^T$.
		\State With the new state $\mathcal{O}_{t+1}$, the agent obtains a reward $r_t$.
		
		\State Update the state $\mathcal{O}_t = \mathcal{O}_{t+1}$.	
%
		
		\EndWhile
		\State $R = 0$
		
		\For{$\tau = t-1:t_{\text{start}}$}
		\State $R \leftarrow r_{\tau} + \lambda * R$
		\State Critic calculates the TD error: $ \delta^{\mu_\theta} = R + \gamma V(\mathcal{O}_{\tau+1}) - V(\mathcal{O}_\tau) $
		\State Update the critic by minimizing the loss: $\mathcal{L}(\theta) = (\delta^{\mu_\theta} )^2$
		\State Update the actor policy by maximizing the action value: $\Delta\theta_{\tau} = \alpha \nabla_{\theta_{\tau}} \log \mu_{\theta_{\tau}}(\mathcal{O}_{\tau}, a_{\tau}) \delta^{\mu_{\theta_{\tau}} }$, $\alpha \in (0,1)$.	
		\EndFor
		
		\State Reveal the true hypothesis, and update samples to the corresponding $\mathcal{F}_{i, j}$, and update the estimated sample density distributions.
		\EndFor
		
		\State Save the trained neural networks.
	\end{algorithmic}
\end{algorithm}

\subsection{Testing Phase}
In the testing phase, the agent first reloads the neural network parameters from the training phase, and makes direct use of the well-trained neural networks without further updates. To test the ability of detecting a change point, we assume that at the beginning of every episode, the hypothesis $H_0$ is true. And to activate the state, $H_0$ will be true for at least $T_1$ time slots so that the agent can learn a high posterior probability of $H_0$. Then, based on the prior belief, a new true hypothesis will be generated, and the agent continues to choose sensors. When the posterior belief of $H_0$ is less than the lower bound $\pi_{\text{low}}$, the agent will report a change point and reset the state to the prior belief. Subsequently, the agent keeps collecting samples until it can claim any of the hypotheses being true.

The full framework is provided in Algorithm \ref{alg:AC-Test} below on the next page.
\begin{algorithm}
	\caption{Deep Actor-Critic Reinforcement Learning Algorithm for Anomaly Detection: Testing Phase}
	\label{alg:AC-Test}
	\begin{algorithmic}
		\State Initialize the critic network $V_{\theta}(\mathcal{O} )$ and the actor $\mu_{\theta}(\mathcal{O})$, and reload the trained parameters $\theta$.
		
		\State The agent initializes the sample space $\Omega^0$, and the subsets $\mathcal{F}_{i,m}$, for $i = 1,\dots, N$ and $m = 0, \dots, M-1$.
		\For{$T = 1 : \text{Maximum episode}$}
		\State Set $H_0$ as the true hypothesis.
		\State The agent fetches the prior belief vector $\pi$ as the initial state.
		\For{$t = 1 : T_1$}
		
		\State With the state $\mathcal{O}_t$, the agent selects one out of the $N$ sensors according to the decision policy $a_t = \mu(\mathcal{O}_t| \theta)$ w.r.t. the current policy.
		\State Agent receives the sample $Y_t$ from the chosen sensor and update the posterior belief vector $\pi^t$.
		\State Agent updates the sample $Y_t$ to the sample space $\Omega^T$.
		
		\EndFor
		
		\State Generate a new hypothesis $H_j$ to be true according to the prior belief $\pi$, and $j \in \{0,1, \dots, M-1\}$.
		\State Set $D = 0$, set $\Gamma = 0$
		\State Set $\mathcal{O}_{T_1}$ as the new state.
		\For{$t^{'} = 1 : \text{Maximum sampling time}$}
		\State With the state $\mathcal{O}_{t^{'}}$, the agent selects one out of the $N$ sensors according to the decision policy $a_{t^{'}} = \mu(\mathcal{O}_{t^{'}}| \theta)$ w.r.t. the current policy.
		\State Agent receives the sample $Y_{t^{'}}$ from the chosen sensor and update the posterior belief vector $\pi^{t^{'}}$.
		\State Agent updates the sample $Y_{t^{'}}$ to the sample space $\Omega^T$.
		\If{$\pi_{0}^{t^{'}} \leq \pi_{\text{low}}$}
		\State Agent rejects the hypothesis $H_0$, and report a change point.
		\State Agent resets the state as $\mathcal{O}_{t^{'}+1}$ as the prior probability $\pi$.
		\EndIf
		\If {$\max(\mathcal{O}_{t^{'}+1}) \geq \pi_{\text{up}}$}
		\State Agent accepts the corresponding hypothesis as the true hypothesis.
		\State \textbf{Break Loop}
		
		\EndIf
		
		\EndFor
		
		\State Reveal the true hypothesis, and update samples to the corresponding $\mathcal{F}_{i, j}$, and update the estimated sample density distributions.
		\EndFor
	\end{algorithmic}
\end{algorithm}

\section{Simulation Results}
\subsection{Experiment Settings}
\subsubsection{Environment}
In our experiments, we set the number of processes as $N = 3$, so that the total number of hypotheses is $M = 8$. The definition of each hypothesis and the distribution of the observations from different sensors under the specific hypothesis being true has been given in Table \ref{table:model} in Section II. Here, we assume that the probabilities of each process being abnormal is $P = [0.2, 0.3, 0.1]$, respectively.

\subsubsection{Actor-Critic Neural Network}
The design of our proposed actor-critic framework is shown in the Table \ref{table:network}. This framework consists of two neural networks. Each neural network includes $3$ layers, and the layers are connected with ReLU activation function. To ensure that the critic network is able to guide the update of the actor network, we assign larger learning rate to the critic network. And in order to maintain a stable and high performance, the learning rates decay over time so that the network parameters will not change rapidly when the neurons are well trained.
\begin{table}[]
\small	
\caption{The settings of actor-critic network}
	\label{table:network}
	\begin{tabular}{|l|c|c|}
		\hline
		& actor                  & critic                                      \\ \hline
		first layer   & 200 neurons + ReLU     & 200 neurons + ReLU                          \\ \hline
		second layer   & 200 neurons + ReLU     & 100 neurons + ReLU                          \\ \hline
		output layer  & N neurons + Softmax    & 1 neuron                                    \\ \hline
		learning rate & 0.0005                 & 0.01                                      \\ \hline
	\end{tabular}
\end{table}

\subsection{Training Phase}
In the training phase, we set the bound $\pi_{\text{up}}$ as $0.8$, and run the procedure shown in Algorithm \ref{alg:AC-Train}. To check the performance of the agent at different training steps, we conduct a validation testing after every $1000$ training steps. The validation set consists of $3$ hypotheses randomly selected from the $M$ hypotheses. We denote the validation set as $\mathcal{H} = \{H_{m^1}, H_{m^2}, H_{m^3}\}$, and in the validation testing, we assign the three chosen hypotheses to be true in the order $ H_{m^1} \rightarrow H_{m^2} \rightarrow H_{m^3} $. In the validation phase, each of the three hypotheses will remain to be true for $200$ sampling steps, and the agent selects the sensor with its current policy, but the network will not be updated. Each time the agent is tested with the validation set,  we record the posterior probabilities of the three hypotheses.

\begin{figure}
	\centering
	\includegraphics[width=1.0\linewidth]{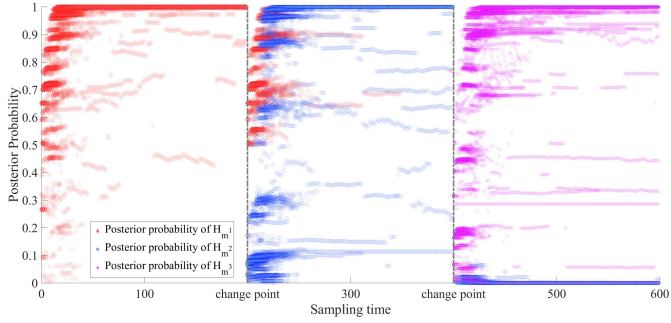}
	\caption{Posterior probability over the sampling time in the validation phase.}
	\label{fig:validation}
\end{figure}

In Fig. \ref{fig:validation}, we plot the posterior probabilities over the sampling time. The posterior probabilities of each hypothesis in the validation set is collected from all validation phases over $15000$ training episodes in total. Since each hypothesis in the validation phase remains to be true for $200$ sampling steps, each validation phase has a fixed duration of $600$ sampling steps. In the figure, the posterior probabilities of different hypotheses are plotted in different colors, and the darkness of the colors stand for the density of the probability at the corresponding value, i.e., the darker the color is, the more frequently that the posterior probability will take the corresponding value at the corresponding sampling time index. We can observe that at the beginning of each change point, the posterior probability of the true hypothesis increases quickly, and remains at a high value that is approximately $1$. And when the next hypothesis starts to be true, the posterior probability of the previous hypothesis diminishes. Besides the patterns with increased darkness, there are also some samples of the probabilities in relatively light colors. The difference in the level of darkness indicates the exploration of the agent while trying to find an efficient selection policy. Since all dark colors appear at high values of the posterior probabilities, the agent is able to detect the true hypothesis with high reliability.

\subsection{Testing Phase}
In the testing phase, we investigate the performance of the proposed agent in terms of the detection delay and loss. Here, we define the claim delay as the difference between the time when the agent claims a hypothesis to be true (i.e., when the posterior probability of the hypothesis exceeds $\pi_{\text{up}}$) and the time when the change occurs. Also, to evaluate the accuracy of the claim, we define the loss as a ratio of the number of wrong claims to the total number of claims. To find a reasonable pair of upper and lower bound for the decision making, in the experiments, we vary the upper bound $\pi_{\text{up}}$ as $\pi_{\text{up}} \in [0.5, 0.55, 0.6, 0.65, 0.7, 0.75, 0.8, 0.85, 0.9, 0.95, 0.99]$, and at the same time vary the lower bound $\pi_{\text{low}}$ as $\pi_{\text{low}} \in [0.1, 0.15, 0.2, 0.25, 0.3, 0.35, 0.4, 0.45, 0.5, 0.55, 0.6]$.

\begin{figure}
	\centering
	\includegraphics[width=1.0\linewidth]{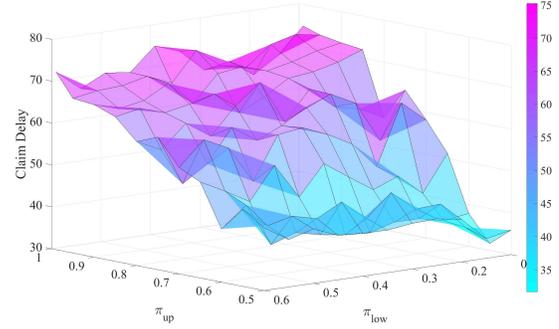}
	\caption{Claim delay under different $<\pi_{\text{up}}$, $\pi_{\text{low}}>$ pairs, when $\pi_{\text{up}} \in [0.5, 0.55, 0.6, 0.65, 0.7, 0.75, 0.8, 0.85, 0.9, 0.95, 0.99]$, and $\pi_{\text{low}} \in [0.1, 0.15, 0.2, 0.25, 0.3, 0.35, 0.4, 0.45, 0.5, 0.55, 0.6]$}
	\label{fig:cdelay}
\end{figure}

\begin{figure}
	\centering
	\includegraphics[width=1.0\linewidth]{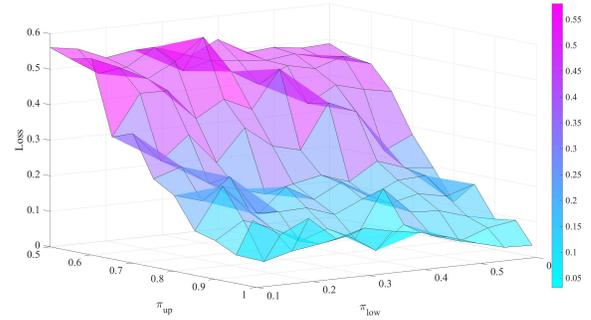}
	\caption{Loss under different $<\pi_{\text{up}}$, $\pi_{\text{low}}>$ pairs, when $\pi_{\text{up}} \in [0.5, 0.55, 0.6, 0.65, 0.7, 0.75, 0.8, 0.85, 0.9, 0.95, 0.99]$, and $\pi_{\text{low}} \in [0.1, 0.15, 0.2, 0.25, 0.3, 0.35, 0.4, 0.45, 0.5, 0.55, 0.6]$}
	\label{fig:loss}
\end{figure}

In Fig. \ref{fig:cdelay} and Fig. \ref{fig:loss}, we plot the average claim delay and average loss, respectively, under each pair of the upper and lower bounds. From the figures we notice that as the upper bound $\pi_{\text{up}}$ increases, the claim delay increases and the loss decreases. This is because when the upper bound is high, the agent accepts a hypothesis more cautiously and hence more observations will be taken, which also improves the confidence level of the decision. On the other hand, as the lower bound $\pi_{\text{low}}$ decreases, the loss also decreases slightly, because the lower bound is the threshold to reject the previous hypothesis that the agent considers to be true (which should always be $H_0$ in the testing phase). When the lower bound is reduced, more stringent conditions are imposed to reject a hypothesis, which results in reduced false alarms. And comparing with the patterns shown in Fig. \ref{fig:validation}, more sampling time is needed in the testing phase. That is because in the testing phase, the detection starts with the posterior probability of $H_0$ being very high, and hence the agent will need more samples to confirm that the previous hypothesis has turned to be false. This ability to adapt to different initializations makes the agent more practically appealing in dealing with the real anomaly detection cases where all processes are normal at the beginning.



\begin{figure}
	\centering
	\includegraphics[width=1.1\linewidth]{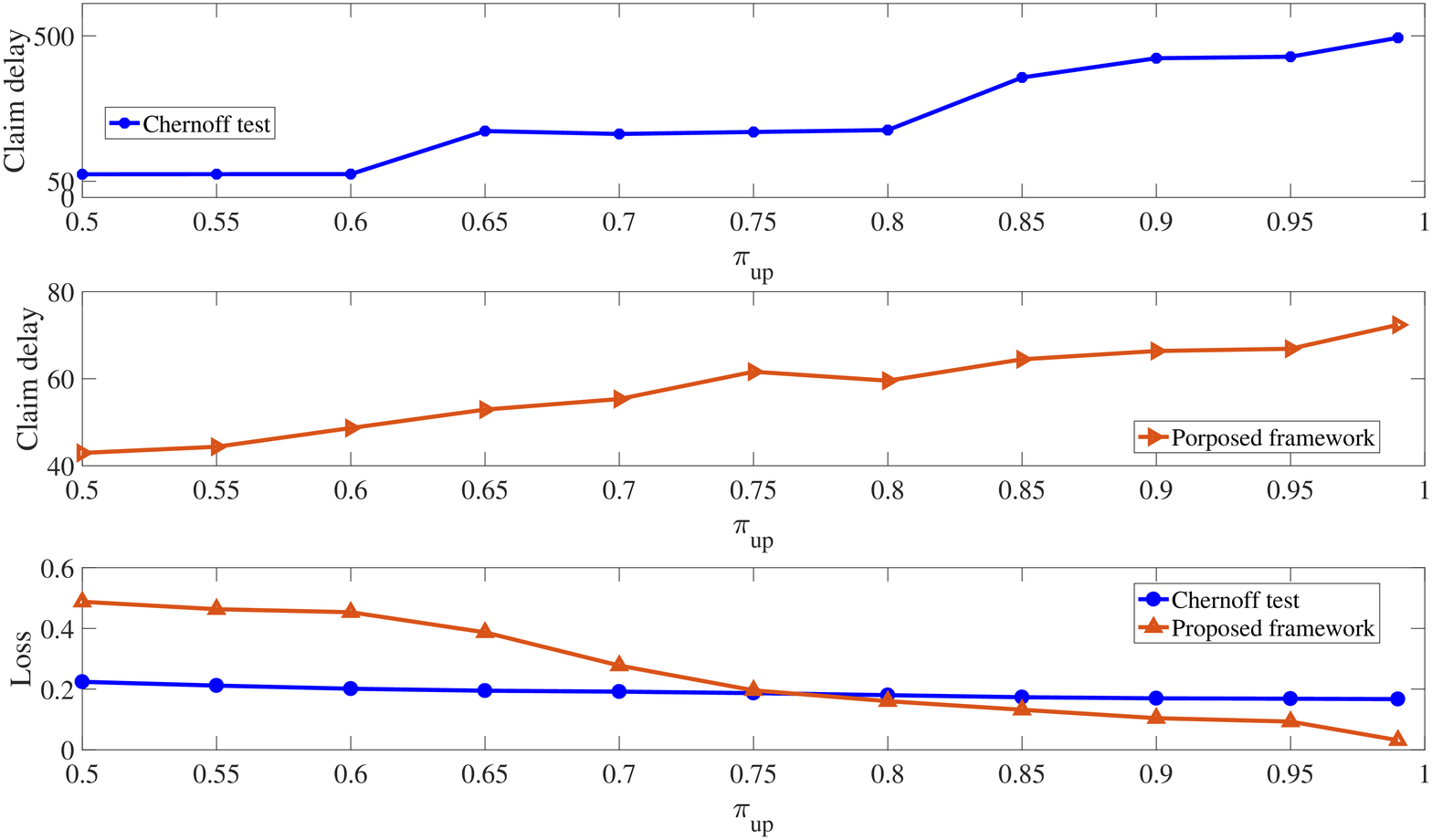}
	\caption{Comparison between proposed framework and Chernoff test: claim delay and loss with $\pi_{\text{low}} = 0.6$, and $\pi_{\text{up}}$ varies as $[0.5, 0.55, 0.6, 0.65, 0.7, 0.75, 0.8, 0.85, 0.9, 0.95, 0.99]$.}
	\label{fig:compare}
\end{figure}

Finally, we compare our proposed framework with the Chernoff test \cite{chernoff1959sequential}. Chernoff test considers the Kullback-Leibler information of the two distributions of the observations, and decides whether to receive the sample from the sensor with highest accumulated log-likelihood ratio or randomly pick one of the sensors. In our experiments, we assign the lower bound $\pi_{\text{low}}$ to be $0.6$, and vary the upper bound as $\pi_{\text{up}} \in [0.5, 0.55, 0.6, 0.65, 0.7, 0.75, 0.8, 0.85, 0.9, 0.95, 0.99]$. Shown in Fig. \ref{fig:compare} are the claim delay and decision loss curves achieved by our proposed framework and Chernoff test. For the claim delay, it is obvious that the Chernoff test will need many more samples to reach the stopping criterion. This is because the Chernoff test assumes that all hypotheses are distinguishable under different tests, which means that it requires all hypotheses to have different observation distributions under each test. However, in our system model, just as shown in Table \ref{table:model}, different hypotheses can have the same observation distribution. For example, under $H_1$ being true, if the agent tests with the sample from sensor $1$, it will not be able to distinguish hypotheses $H_1$, $H_4$, $H_5$, and $H_7$, because under all these hypotheses, the process $1$ is in abnormal state. And for the loss, it is obvious that the loss from the proposed agent decreases when the upper bound increases. However, the loss from the Chernoff test, though slightly decreases as the upper bound gets larger, is relatively stable. When $\pi_{\text{up}} \geq 0.75$, the performance of the proposed agent is more competitive in terms of both the claim delay and loss. So the proposed agent is more suitable for systems with high sampling cost and require high confidence levels.

\section{Conclusion}
In this work, we have considered active sequential testing for anomaly detection, in which an unknown number of processes could be in abnormal states simultaneously. To solve the dynamic problem of how to select sensors based on a partially observable Markov decision process, we have proposed a deep actor-critic reinforcement learning framework, which enables the agent to dynamically select the sensors and minimize the claim delay while maximizing the confidence level based on the posterior probabilities. We have designed the actor-critic sensor selection algorithm, refining the updating procedure. We have analyzed the performance of the proposed framework. In particular, in the training phase, we have conducted validation testing and demonstrated the convergence of the posterior probabilities. In the testing phase, we have investigated the selection of upper and lower thresholds and their influence on the claim delay and loss. Finally, we have provided comparisons between the proposed framework and Chernoff test, and demonstrate the superior performance of the proposed actor-critic deep reinforcement learning framework in terms of lower claim delay. Additionally, while the Chernoff test has lower loss for smaller values of upper threshold, the proposed framework outperforms when higher confidence levels are required (i.e., for larger values of the upper threshold).

\bibliographystyle{ieeetr}
\bibliography{reference}

\end{document}